# Superposed Episodic and Semantic Memory via Sparse Distributed Representation

**Rod Rinkus**
Neurithmic Systems
468 Waltham St.
Newton, MA 02465
*rod@neurithmicsystems.com*

**Jasmin Leveille**
Scientific Systems Co Inc
500 W Cummings Park #3000
Woburn, MA 01801
*jalev51@gmail.com*

## Abstract

The abilities to perceive, learn, and use generalities, similarities, classes, i.e., "semantic memory" (SM), is central to cognition. Machine learning (ML), neural network, and AI research has been primarily driven by tasks requiring such abilities. However, another central facet of cognition, the ability to rapidly form, often with a single occurrence, essentially permanent memories of experiences, i.e., "episodic memory" (EM), has received relatively little attention. Only recently has EM-like functionality begun to be incorporated into Deep Learning (DL) models, e.g., Neural Turing Machines, Memory Networks. However, in these cases: a) EM is implemented as a separate module, which entails substantial data movement (and thus, time and power) between the DL network itself and EM; and b) the individual items are stored localistically within the EM, precluding realizing the exponential representational efficiency of distributed over localist coding. We describe Sparsey, an unsupervised, hierarchical, spatial/spatiotemporal associative memory model differing fundamentally from mainstream ML approaches, most crucially, in that it represents inputs in the form of *sparse distributed representations* (SDRs), or, *cell assemblies*, which admits an extremely efficient, single-trial learning algorithm that maps input similarity into code space similarity (measured as intersection). SDRs of individual inputs are stored in superposition and because similarity is preserved, the patterns of intersections over the assigned codes reflect the similarity, i.e., statistical, structure, of all orders, not simply pairwise, over the inputs. Thus, SM, i.e., a generative model, is built as a computationally free side effect of the act of storing episodic memory traces of individual inputs, either spatial patterns or sequences. We report initial results on MNIST and on the Weizmann video event recognition benchmarks. While we have not yet attained SOTA class accuracy, learning takes only minutes on a single CPU.

## 1 Introduction and Model Overview

Intelligence is strongly associated with the ability to learn generalities and distinctions in the world, which we can term, *semantic memory* (SM), and mechanizing such abilities has for decades been the primary driver of machine learning (ML), neural network, and AI research. In contrast, another essential facet of cognition, the ability to form essentially permanent memories of individual experiences, called *episodic memory* (EM), has received far less attention. Only recently has EM-like functionality begun to be incorporated into SOTA Deep Learning (DL) models [1-4], which has been shown to increase data efficiency and enable rapid learning of long time lag dependencies, e.g., in the context of reinforcement learning [4]. However, in such models, EM has been added to the overall model structure as a separate module. This means that significant computational time and power must be devoted to moving information back and forth between the primary network, e.g., a DL network, which interfaces with the environment, and the EM. Furthermore, the EM in such models represents (stores) the individual episodic traces physically separately from each other, i.e., *localistically*, which precludes realizing the exponential representational advantage of distributed representation—e.g., a field of *N* binary units can represent $2^N$ codes, which is exponentially greater than the *N* codes possible under localism.

We describe Sparsey, an unsupervised, hierarchical, spatial/spatiotemporal associative memory model that addresses both the above drawbacks. Sparsey differs fundamentally from mainstream ML/DL approaches, most crucially, in that it represents inputs in the form of *sparse distributed representations* (SDRs), or, *cell assemblies* [5], as opposed to either dense/fully distributed or localist representations. An overall Sparsey instance is a hierarchy of SDR coding fields, called "macs" because we've proposed such a field as a model of the L2/3 portion of a cortical macrocolumn [6]. An input, e.g., a frame of video, is mapped to a pattern of activation composed of SDRs (codes) in multiple of the macs throughout the hierarchy, as in Figure 1c. Codes (more precisely, the active cells comprising the codes) become linked using single-trial, Hebbian-type learning, within levels [via horizontal (H) synaptic matrices] and across levels [via bottom-up (U) and top-down (D) matrices], resulting in hierarchical, spatiotemporal memory traces of input sequences. Further, codes at higher levels have longer activation durations (*persistences*), thus higher-level codes associate with sequences of lower-level codes (i.e., chunking, compression).

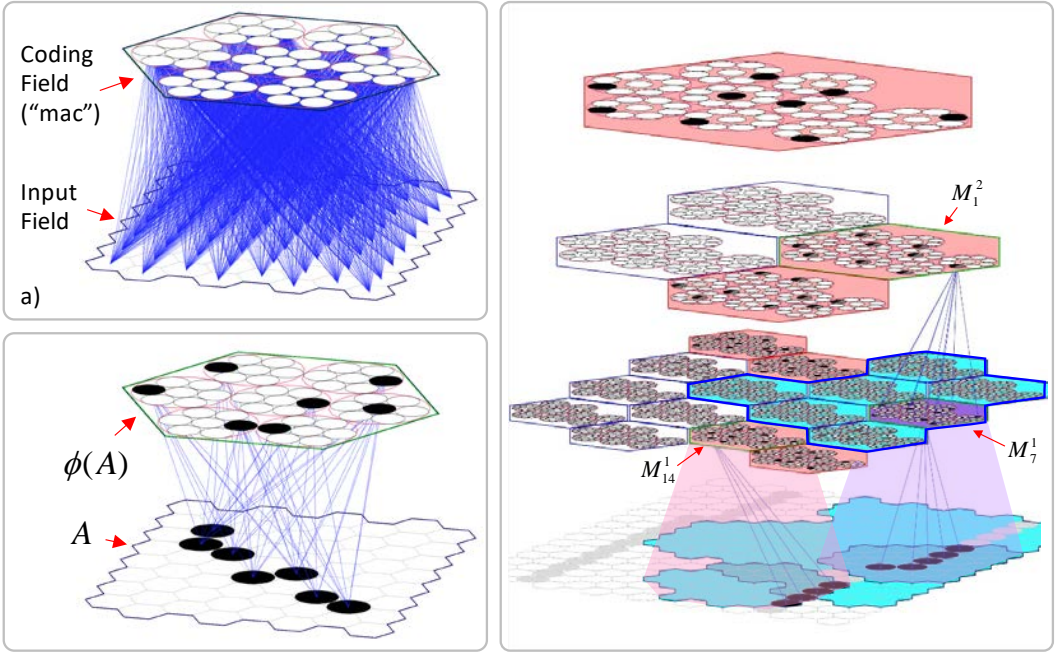

Figure 1: a) An SDR coding field ("mac") consisting of $Q=7$ WTA competitive modules (CMs), each with $K=7$ units. b) An input A, its code, $\phi(A)$, consisting of $Q$ active (black) units, and the weights increased to form the association. c) Hierarchical model instance showing subset of active macs across all levels for a single input frame of a sequence. The large cyan patch at L0 is the combined L0 receptive field (RF) of the seven L1 macs comprising $M_1^2$'s immediate (L1) RF.

The heart of Sparsey is its *Code Selection Algorithm* (CSA), for activating a mac code in response to its total input (which generally includes U, H, and D inputs) during learning and retrieval. See [6-8] for detailed CSA description. To briefly summarize, the CSA computes the familiarity *G* of the total, i.e., context-dependent, input to a mac and activates an SDR whose intersections with the SDRs of similar stored inputs increases with *G*. As *G* goes to zero (complete unfamiliarity), the expected intersection of the activated SDR with previously stored SDRs goes to chance. Thus, the CSA preserves similarity, moving by degrees between code completion in proportion to the input's familiarity and code separation in proportion to its novelty. This implies that the similarity (statistical, correlational) structure in the input space is reflected in the patterns of intersections over the SDRs stored in a mac. Thus, while Sparsey's primary activity is on-line, single-trial creation of memory traces of detailed memory traces of inputs, i.e., EM, SM automatically emerges as a computationally free side-effect. Both the EM traces and SM (higher-order intersection patterns over EM traces) are stored in superposition; hence, no communication cost between EM and SM.

A mac consists of $Q$ WTA competitive modules (CMs), each consisting of $K$ units; Figure 1a shows a small mac instance with its input field, or receptive field (RF), of binary pixels, which are completely connected to the mac's units with low precision weights, all initially zero. Figure 1b shows a mac code (denoted by $\phi$), consisting of one winner per CM. We emphasize that the sparseness of the code is imposed *structurally*: no computation is needed explicitly for imposing/managing sparseness as is the case in the far more ubiquitous "sparse coding" models [9], wherein the sparsity penalty is part of the objective function which must be repeatedly evaluated. Figure 1c shows several key model properties: a) internal levels generally have an explicit mesoscale, e.g., L1 is a 4x4 array of macs; b) RFs of neighboring macs can overlap; c) the unit's comprising RFs of higher level macs are whole macs, e.g., $M_1^2$'s RF (heavy blue outline in L1) consists of seven L1 macs, one of which, $M_7^1$, is active; and d) generally, only a fraction of the macs are active for any given input, providing a further source of overall sparseness (beyond the *structurally imposed* sparseness of the mac).

## 2 Results

Our results explicitly demonstrate an ability to learn the statistical, specifically the class, structure of the domains, i.e., SM. However, they only partially demonstrate EM. That is, by definition, an episodic memory is formed on the basis of a single trial. Our studies meet that criteria. However, to fully demonstrate EM, we must show that we can prompt with only a portion of a training input and have the model return the remainder, i.e., episodic *recall*. Unfortunately, our simulation was not set up to produce episodic recall data for these models. However: i) we have reported episodic recall evidence for difficult synthetic sequence data sets in earlier work [7]; and ii) we conducted sanity tests, i.e., where the test and train sets are the same, on the reported models, and generally found very high similarity, e.g., 95% or more, of the unit activations are the same, between the memory trace activated at test and on the single training trial (data not reported). Technically, this demonstrates strong episodic *recognition* (as opposed to recall). Note that these hierarchical traces involve hundreds to thousands of unit activations across many macs and across time steps. Future studies will include direct episodic recall tests of the same models used for classification.

### 2.1 MNIST

We preprocessed MNIST data [10] with edge filtering, binarization and skeletonization, resulting in 16x24 binary pixel images such as the instance of "5" shown (in perspective, gray/black pixels active) in the input level of Figure 2. The model had one internal level (L1) consisting of 21×32=672 macs. Active macs are highlighted (red) and a sample of their RFs are highlighted (cyan), with representative U wts (blue). RFs were highly overlapped: neighboring macs shared up to 90% of their pixels. RF size (in pixels) and the numbers of pixels required to activate the macs differed across macs. Consequently, every active pixel generally participates in activating several of the overlying macs and is thus represented multiple times and in multiple contexts at L1, a kind of inter-mac *overcompleteness* which is in addition to the intra-mac overcompleteness present amongst the set of inputs (i.e., basis elements) stored in superposition, as SDR codes, in each mac.

The first series of experiments (Table 1, rows 1-4) involved a subset containing 9,000 MNIST samples (900 per class). The net was trained on 200 samples per class: the test set was chosen randomly from the remaining 700 per class. Experiments involved tests of from 100 to 700 examples per class. The learning that occurred between L0 and L1 was unsupervised. However, a supervised protocol was imposed by having another localist field of 10 binary class-representing units (not shown) that received full top-down (D) connectivity from all cells of L2. For each training instance, the D wts from all active L2 cells to the active class cell were set to binary "1". On each test trial, the input was presented, which led to activation of codes in L2 macs, which then sent outputs to the class field, whereupon the class cell with the maximum summation won. The results (rows 1-4) show that despite the small training set and single-trial learning, reasonably high accuracy of about 90% is achieved for all test set sizes. A second study, experiments 5-9, using a training set of 800 samples per class, and testing on up to 5,000 per class, showed that accuracy held steady at 89% for all test set sizes, revealing very strong generalization from the still relatively small training set.

While we have not yet attained SOA accuracy, which is >99.9%, it is nonetheless high, i.e., chance=10%. Far more important however is that *without using any machine parallelism (MP)*, e.g., GPUs, training time for 200 samples/class is only 220 secs. That model instance had macs with $Q=11$ and $K=9$, for a total of 66,528 L2 cells and ~1,5 million U wts. For the second study (800 samples/class), macs had $Q=16$ and $K=11$, for a total 118,272 L2 cells and ~3.7 million U wts, and training time was 25 minutes.

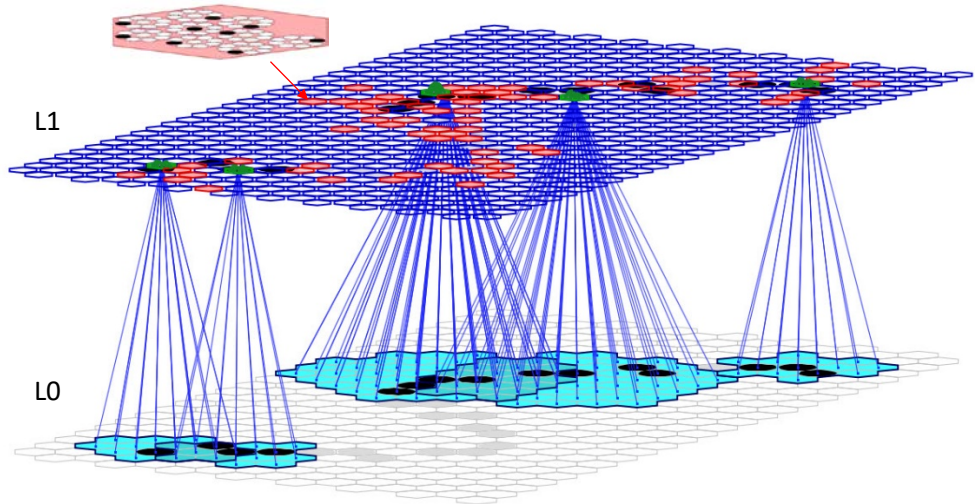

Figure 2: A preprocessed "5" from MNIST is active at L0. L1 is a 21×32 array of macs, 77 of which are active because the numbers of active pixels in their respective RFs are within acceptable ranges. At top, a notional zoom-in of an active mac (actual $Q$ and $K$ values varied across studies).

Table 1: MNIST Studies

| Exp. | Train Sample / Class | Test Samples / Class | Accuracy (%) |
|---|---|---|---|
| 1 | 200 | 100 | 91 |
| 2 | 200 | 200 | 89 |
| 3 | 200 | 400 | 88 |
| 4 | 200 | 700 | 87 |
| 5 | 800 | 500 | 89 |
| 6 | 800 | 1000 | 89 |
| 7 | 800 | 2000 | 89 |
| 8 | 800 | 3000 | 89 |
| 9 | 800 | 5000 | 89 |

While Sparsey's MNIST training times are already short we emphasize that the model is written in java and that substantial speedup is likely easily attainable with code optimization, even without incorporating MP. Furthermore, the parameter space is huge and still largely unexplored. For example, we expect that models with more levels can capture a more parsimonious hierarchical decomposition of the MNIST space (cf., deeper models can more efficiently represent certain nonlinearities [11]), allowing a smaller total number of macs across all levels, smaller $Q$ and $K$ values, thus, fewer total weights, possibly fewer active macs per level on average, and ultimately faster learning (and inference) times. Regarding MP, Initial studies simulating ASIC and processor-in-memory hardware realizations of Sparsey conservatively find 2-3 orders of magnitude speedup [12, 13]. Ideally, we would like to implement Sparsey on a Memristor-like platform given its simple logic, very low-precision activations and weights, and high fault tolerance.

### 2.2 Weizmann Video Event Recognition

The Weizmann video dataset [14] consists of 90 snippets: nine instances of each of 10 event classes, e.g., walking, jumping, waving, etc., performed by nine human actors. We cropped the original frames to 84×120 bounding boxes around the actor, then lowered resolution to 42×60 pixels. We then edge-filtered, binarized, and skeletonized individual frames, and decimated in time, yielding an average of ~10 frames per snippet. Figure 3 (left) shows one of the resulting snippets, decimated to 12 frames. We then created five additional, noisy versions of each of the 90 preprocessed snippets (on each frame, randomly "moving" 20% of the "on" pixels to nearby "off" locations contiguous to an "on" pixel, as in Fig. 3), resulting in a total *augmented* dataset of 540 videos.

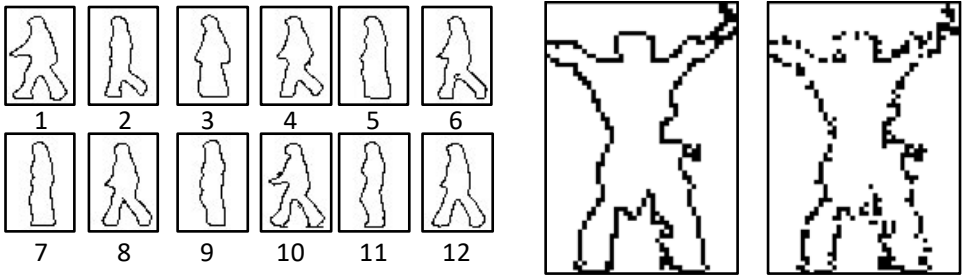

Figure 3: (left) One of the Weizmann "walk" snippets, preprocessed as described in the text. (right) Example of a preprocessed frame and a 20% noisy variant.

The learning protocol had two phases. In the first, unsupervised phase, the 540 snippets were presented once each to the model of Figure 4, which had 12×18=216 L1 macs and 6×9=54 L2 macs and ~6.2 million wts. The L2 macs had $Q$=6 and $K$=6 yielding a total of 54×36=1,944 binary units. The total L2 activation pattern present on the last frame of a snippet (typically having ~15-20 active macs whose codes depend on the temporal context of the entire sequence) is taken as the code for the snippet as a whole. This training time was ~210 secs. We then used these 540 1,944-bit vectors in a leave-one-out (LOO) supervised SVM training protocol. Specifically, for each of the nine actors, training an SVM on the snippet-final vectors of the other eight actors' original and noisy snippets (a total of 480 vectors) and testing on the 10 original snippets for the withheld actor. Using this protocol, 60 of the 90 test snippets were correctly recognized, i.e., 67% accuracy. While this is well below SOA (which is 100%), it is well above chance (10%). But again, the more important result is the very short training time achieved without using MP. We are currently exploring the large parameter space to find deeper, but smaller and thus faster, models that achieve accuracies much closer to SOA.

We emphasize that the inputs are simply binary pixels. We have not experimented with using more informative features as one of our major objectives is to explain how the brain learns everything from scratch, starting with the simplest features, edge-like, possibly moving/morphing, contours in small L1 (analogous to cortical V1) mac RFs, e.g., comprised of ~40-100 pixels, and building up through progressively larger RFs within which progressively more complex spatiotemporal feature compositions occur. For example, in Figure 4, mac $M_{15}^2$'s L0 RF covers a large portion of the input surface, within which a huge space of higher-level objects/events could occur. The input on this frame results in 33 L1 macs activating which then cause 17 L2 codes to activate, for a total of 17×6=102 active L2 cells (out of 1,944). The L1 RF of $M_{15}^2$ consists of 35 L1 macs, five of which are active (purple). $M_{15}^2$'s L0 RF is the union of the L0 RFs of those 35 L1 macs, two of which are highlighted. Black pixels are those causing those two associated L1 macs to activate. We can consider $M_{15}^2$'s U input to represent five features, or parts, one represented in each of its five active L1 macs. In general, those parts may overlap at the pixel level (e.g., as in Figure 2). Although the L1 cells are not shown here, the general scenario is that during learning a set of SDR codes will be stored in each mac, representing a basis of features which will be idiosyncratic to the mac's particular history of inputs. Thus, some of the information about the contents of $M_{15}^2$'s L0 RF is carried by the particular code activated in each of its active L1 macs, and some is carried by the spatial positions of those active L1 macs within $M_{15}^2$'s L1 RF. This, in combination with the fact that the macs are essentially autonomous and each mac learns the statistical, and thus, invariance, structure of its particular spatiotemporal aperture onto the world, is consistent with the emerging notion of staged disentangling of inputs across a hierarchy [15-17].

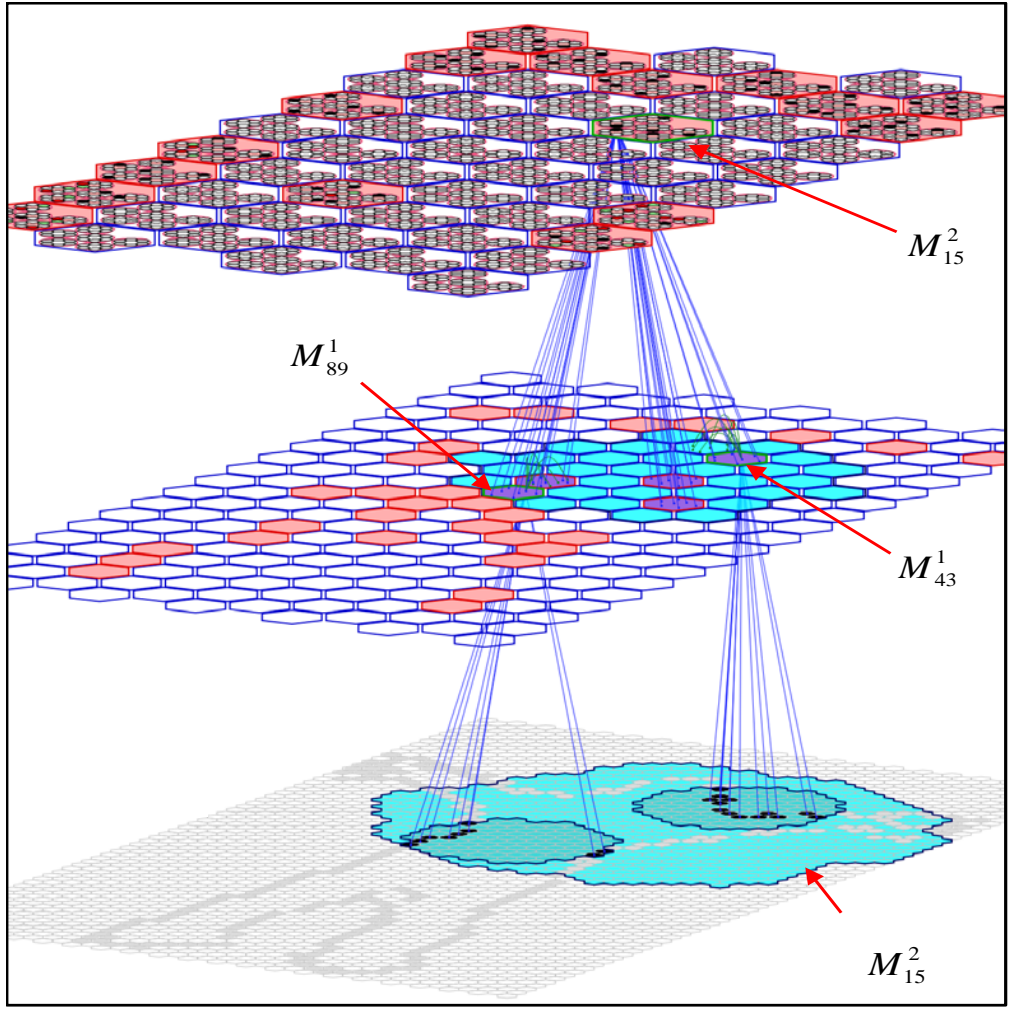

Figure 4: 3-level model used for Weizmann data, highlighting nested receptive field (RF) structure.

## 3  Discussion

Deep learning methods, i.e., Deep Belief Nets, ConvNets, LSTM, have achieved great successes in recent years, significantly lowering error rates on numerous ML benchmarks, and learning complex mappings, i.e., action policies, e.g., AlphaGo. The incorporation of episodic memory (EM) functionality is just one of several advances that have been crucial, including contrastive divergence [18], level-by-level unsupervised pretraining, dropout, tied/shared weights, incorporating reinforcement learning (RL) [19], adversarial learning. However, the two items usually mentioned first as fueling the rise and success of DL are: 1) availability of cheap, massive *machine parallelism* (MP), i.e., GPUs; and 2) availability of massive amounts of training data. There are two major causes for concern regarding item 1. First, it is readily acknowledged that without massive MP, DL learning times are unacceptably long. Thus, the current situation for DL is that learning either requires lots of power or takes too long. Secondly, though the brain is massively parallel, the forms of parallelism used in DL, model and data parallelism, and tied/shared weights, are clearly non-biological in nature, and actually *accentuate* the processor-memory distinction despite the fact that most (~80-90%) of the energy used in computation is expended in moving data. And, as noted at the outset, the physical separation of the primary model, i.e., the locus of semantic memory (SM), and EM also incurs increased computational costs. The second item above, availability of massive training data, is also problematic, because it engenders a view of learning that may be quite at odds with human learning. It is increasingly acknowledged that much of learning, particularly of declarative knowledge, appears to be single or few-trial. Certainly, by definition, *episodic* memories of specific events are formed on the basis of single trials. These issues raise questions as to whether DL learning algorithms and representations might fundamentally differ from those of the brain.

We've described a model, Sparsey, which differs fundamentally from mainstream SOTA DL approaches and we provided results demonstrating that it learns higher-order statistical structure of the input space—evidenced by good (though not yet SOTA) classification performance on spatial/spatiotemporal data—as a side effect of storing SDR memory traces of individual inputs in superposition based on single trials. This simultaneously addresses both weaknesses of current mainstream EM treatments identified at the outset. That is, since SM and EM are co-located in superposition, there is: a) no need to communicate data between SM and EM; and b) the efficiency of distributed representation, specifically of SDR, is leveraged by both SM and EM. In fact, it is this leveraging of SDR's representational efficiency combined with single-trial learning that underlies Sparsey's extremely fast, especially considering that no machine parallelism (MP) is used, learning times. Given the easily achievable 2-3 orders of magnitude speedup that MP will provide [12, 13], we believe our results clearly show that Sparsey outperforms mainstream DL approaches in terms of learning speed. In fact, as we have previously described [6-8], both storing (learning) a new item in a mac and retrieving the best-matching item from a mac are f*ixed time* operations, i.e., the number of operations needed to do either remains constant as the number of items stored in a mac increases. And, since a storage or retrieval operation for an overall model (i.e., hierarchy of macs) involves a single iteration over its macs, the overall model also has fixed-time performance. This capability is essential for scaling to truly massive data sizes and to our knowledge, has not been shown for any other model, including any hashing model [20, 21]; see [22] for a review. While several SDR-based models have been put forth [23-26], we believe Sparsey is the first to demonstrate SDR in conjunction with hierarchy, an explicit mesoscale (tiling of macs) at each hierarchical level, and sequence memory via recurrence.

Stepping back, we can ask: is the brain's fundamental purpose to remember inputs / events that an organism actually experiences, i.e., EM, or to learn the class structure of the world, i.e., SM? Clearly, the vast majority of ML research has focused on classification and learning statistical (generative) models of domains, with relatively little concern for being able to recall, in full detail, individual experiences that may have occurred remotely in a model's operational life. In fact, *catastrophic forgetting*, in which new learning erases old learning has been a perennial problem for ML/DL, right up to the present, as discussed in [27], which presents a novel solution. It is perhaps tempting to view the neocortex-hippocampus distinction as supporting the idea that SM and EM functionality are physically separated. That is, substantial evidence suggests that the hippocampus forms immediate memory traces of experiences, replays the traces (e.g., during sleep), which gradually embeds (consolidates) permanent traces in neocortex, and presumably allows those immediately formed hippocampal traces to fade or be overwritten, evidenced by the fact that old knowledge can be accessed without the hippocampus. However, mounting evidence supports the idea that the final repository not only of semantic memories but also of episodic memories is neocortex. In this case, a theory like Sparsey might provide a more parsimonious explanation of their co-location, and one which has significant computational advantages.

**Acknowledgement**

We thank the Office of Naval Research and Darpa for supporting this research in recent years and colleagues, Greg Lesher, Oliver Layton, Harald Ruda, and Nick Nowak for help with this research.